\documentclass[dvipsnames,table]{article}
\usepackage{colm2024_conference}
\pdfoutput=1

\usepackage{amsmath,amsfonts,bm}

\def\eqref#1{equation~\ref{#1}}

\def\1{\bm{1}}

\DeclareMathAlphabet{\mathsfit}{\encodingdefault}{\sfdefault}{m}{sl}
\SetMathAlphabet{\mathsfit}{bold}{\encodingdefault}{\sfdefault}{bx}{n}

\usepackage[utf8]{inputenc}
\usepackage[T1]{fontenc}
\usepackage{textcomp}
\usepackage{microtype}
\usepackage{lipsum}
\usepackage{amsmath}
\usepackage{amsfonts}
\usepackage{amssymb}
\usepackage{amsthm}
\usepackage{nicefrac}
\usepackage{xcolor}         
\definecolor{scaleBlack}{HTML}{000000}
\definecolor{scaleWhite}{HTML}{FFFFFF}
\definecolor{scaleInk}{HTML}{2E2E2E}
\definecolor{scaleDarkGray}{HTML}{2E2E2E}
\definecolor{scaleGray}{HTML}{7D7D7D}
\definecolor{scaleMediumGray}{HTML}{C7C7C7}
\definecolor{scaleLightGray}{HTML}{EAEAEA}
\definecolor{scalePanel}{HTML}{F7F7F7}
\definecolor{scaleLink}{HTML}{598DD2}

\definecolor{scaleTerracotta}{HTML}{E27553}
\definecolor{scaleTeal}{HTML}{65C2B9}
\definecolor{scaleBlue}{HTML}{598DD2}
\definecolor{scaleGold}{HTML}{D1B745}
\definecolor{scalePurple}{HTML}{A37CAD}
\definecolor{scalePink}{HTML}{E072AF}

\definecolor{scaleAmber}{HTML}{E09A3B}
\definecolor{scaleSky}{HTML}{6FCBEF}
\definecolor{scaleGreen}{HTML}{79C96E}
\definecolor{scalePeriwinkle}{HTML}{8D86E8}
\definecolor{scaleCoral}{HTML}{D95358}
\definecolor{scaleSalmon}{HTML}{E39CAB}
\definecolor{scaleYellow}{HTML}{EDE272}

\definecolor{scaleTan}{HTML}{BD836F}
\definecolor{scaleOlive}{HTML}{A69E66}
\definecolor{scaleForest}{HTML}{62A16E}
\definecolor{scaleSlate}{HTML}{75A4B5}
\definecolor{scaleMauveDark}{HTML}{817EAD}
\definecolor{scaleDustyRose}{HTML}{D97D7C}

\definecolor{scalePeach}{HTML}{EDCCA3}
\definecolor{scaleSand}{HTML}{C7A678}
\definecolor{scaleSage}{HTML}{9AB896}
\definecolor{scalePowderBlue}{HTML}{8FB6EB}
\definecolor{scaleMauve}{HTML}{D1B4C6}
\definecolor{scaleLavender}{HTML}{DDB5EB}

\definecolor{scalePositive}{HTML}{79C96E}
\definecolor{scaleNeutral}{HTML}{EDE272}
\definecolor{scaleNegative}{HTML}{D95358}

\colorlet{scaleTerracottaFill}{scaleTerracotta!82!white}
\colorlet{scaleTealFill}{scaleTeal!82!white}
\colorlet{scaleBlueFill}{scaleBlue!82!white}
\colorlet{scaleGoldFill}{scaleGold!82!white}
\colorlet{scalePurpleFill}{scalePurple!82!white}
\colorlet{scalePinkFill}{scalePink!82!white}
\colorlet{scaleTanFill}{scaleTan!82!white}
\colorlet{scaleSlateFill}{scaleSlate!82!white}

\colorlet{scaleEvergreen}{scaleTerracotta}
\colorlet{scaleEvergreenLine}{scaleTerracotta}
\colorlet{scaleEvergreenFill}{scaleTerracottaFill}
\colorlet{scaleAtlas}{scaleTeal}
\colorlet{scaleAtlasLine}{scaleTeal}
\colorlet{scaleAtlasFill}{scaleTealFill}
\colorlet{scaleTanLine}{scaleTan}
\colorlet{scalePurpleLine}{scalePurple}
\colorlet{scaleSlateLine}{scaleSlate}

\definecolor{darkgrey}{rgb}{0.53,0.53,0.53}
\definecolor{mygrey}{rgb}{0.9,0.9,0.9}
\definecolor{tabhighlight}{HTML}{e5e5e5}
\definecolor{prussianblue}{RGB}{0,51,102}
\definecolor{fbApp}{HTML}{ffe4e3}
\definecolor{lightgray}{gray}{0.9}

\definecolor{scaleWine}{HTML}{7D2935}   
\definecolor{scaleCrimson}{HTML}{C0202C} 
\definecolor{scaleCardinal}{HTML}{C8102E} 
\definecolor{scaleVermilion}{HTML}{E23A2E}
\usepackage{booktabs}
\usepackage{array}
\usepackage{multirow}
\usepackage{makecell}
\usepackage{tabularx}
\usepackage{tabulary}
\usepackage{colortbl}
\usepackage{graphicx}
\usepackage{float}
\usepackage{wrapfig}
\usepackage{caption}
\usepackage{subcaption}
\usepackage{placeins}       
\usepackage{pifont}
\usepackage{siunitx}
\usepackage{enumitem}
\usepackage{comment}
\usepackage[normalem]{ulem}
\usepackage{soul}
\usepackage{mdframed}
\usepackage[many]{tcolorbox} 
\usepackage{algorithm}
\usepackage{algorithmic}
\usepackage{tikz}
\usetikzlibrary{positioning,fit,backgrounds,calc,svg.path,shapes.geometric,arrows}

\usepackage{url}

\definecolor{growGreen}{HTML}{1B7F4A}

\newlength{\rlvrTitleW}
\newcommand{\rlvrTitleSecond}{%
  \settowidth{\rlvrTitleW}{\hspace{2.5em}RLVR$^{2}$: Reinforcement Learning with Verifiable}%
  \makebox[\rlvrTitleW][c]{Rubric-based Ranking}}
\title{RLVR$^{2}$: Reinforcement Learning with Verifiable\\\rlvrTitleSecond}

\author{%
\makebox[\linewidth][c]{%
\begin{tabular}{@{}c@{}}
\bf Hao Li$^{*}$$^{\ddagger}$, \quad Zhengkun Zhang$^{*\dagger}$, \quad Gangqiang Hu, \quad Zhen Zhang \\
\bf Yude Gao, \quad Dai Dai, \quad Jing Liu \\[0.3em]
\textbf{ERNIE Team, Baidu}
\end{tabular}}%
}

\usepackage{etoolbox}
\newcommand{\rlvrdate}{Nov 5, 2026}
\patchcmd{\mymaketitle}{Feb 4, 2026}{\rlvrdate}{}{\PackageWarning{rlvr}{date patch failed}}
\makeatletter
\fancypagestyle{normalpage}{
  \fancyhf{}
  \lhead{\includegraphics[height=15pt]{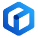}}
  \chead{\textcolor[HTML]{4D4D4D}{\textbf{\thetitle}}}
  \rhead{\textcolor[HTML]{4D4D4D}{\textbf{\rlvrdate}}}
  \renewcommand{\headrulewidth}{0.4pt}
  \renewcommand{\headrule}{{\color[HTML]{4D4D4D}\hrule\@height\headrulewidth\@width\headwidth\vskip-\headrulewidth}}
  \cfoot{\thepage}
}
\makeatother

\makeatletter\def\@abstract{
Reinforcement Learning with Verifiable Rewards (RLVR) is expanding from tasks with relatively well-defined correctness signals, such as mathematics and code, toward multifaceted quality requirements, where quality is typically specified by multi-dimensional rubrics rather than a single criterion. Since policy optimization generally consumes one scalar per rollout, rubric-based RL pipelines must map multiple criterion scores into a scalar reward. This aggregation step is often treated as score scaling, although it implicitly determines how different quality dimensions trade off during training. A prevailing practice normalizes each criterion and takes a linear aggregation, which implicitly assumes that cardinal score differences are comparable across criteria and that gains on one criterion can compensate for failures on another---assumptions that are unreliable when rubric criteria are semantically heterogeneous. We propose \textbf{Reinforcement Learning with Verifiable Rubric-based Ranking} (\textbf{RLVR$^{2}$}), a verifiable ranking paradigm for rubric-based RLVR. For each criterion, RLVR$^{2}$ converts rubric scores into criterion-specific within-group ordinal outcomes and recovers a latent utility from the resulting within-group comparison matrix;  these utilities are then converted into a single optimization signal for policy training. By retaining only within-group ordering and discarding raw score magnitudes, RLVR$^{2}$ avoids calibrating heterogeneous rubric scales. RLVR$^{2}$ further supports objective-preserving attribute adjustment: auxiliary attributes that are systematically associated with observed rankings but are not themselves training objectives can be incorporated into the estimation process without expanding the rubric or directly rewarding those attributes. Across three model scales and 20 benchmarks, RLVR$^{2}$ consistently outperforms a broad range of representative rubric-based RLVR baselines, achieves the best overall performance on a majority of benchmarks at every scale. Further analysis shows that it controls systematic effects associated with reasoning efficiency and response formatting while preserving the underlying quality objective.

}\makeatother

\let\rlvrOldSection\section
\renewcommand{\section}{\FloatBarrier\rlvrOldSection}

\begin{document}
\begingroup
\renewcommand\thefootnote{}
\footnotetext{%
  $^{*}$Equal Contribution. \quad $^{\ddagger}$Project Lead. \quad $^{\dagger}$Corresponding Author.\newline
  \hspace*{2em}Contact email: \href{mailto:haoli47@outlook.com}{h.li1@imperial.ac.uk} or 
  \href{mailto:haoli47@outlook.com}{haoli47@outlook.com},
  \href{mailto:zhangzhengkun2023@gmail.com}{zhangzhengkun2023@gmail.com}%
}
\endgroup
\maketitle
\renewcommand{\thetitle}{Reinforcement Learning with Verifiable Rubric-based Ranking}

\pagestyle{normalpage}
\clearpage

\section{Introduction}
\label{sec:intro}




Reinforcement learning with verifiable rewards (RLVR) has become a central paradigm for post-training large language models \citep{DBLP:journals/corr/abs-2501-12948}. As the qualities expected of tasks grow increasingly rich and multi-dimensional, a single holistic reward can no longer capture them \citep{DBLP:journals/corr/abs-2508-12790, DBLP:journals/corr/abs-2605-26579}. A question-answering response, for instance, may need to follow instructions, stay factually correct, remain in the query's language, and stay safe \citep{DBLP:conf/acl/JiaZCLJQ26}. Across this task spectrum, per-criterion scores must ultimately be fused into a single scalar reward for existing policy-optimization algorithms \citep{schulman2017proximal, DBLP:journals/corr/abs-2402-03300}. This raises the central question of how heterogeneous rubric criteria should be aggregated.

\begin{figure*}[t]
    \centering
    \includegraphics[width=\textwidth]{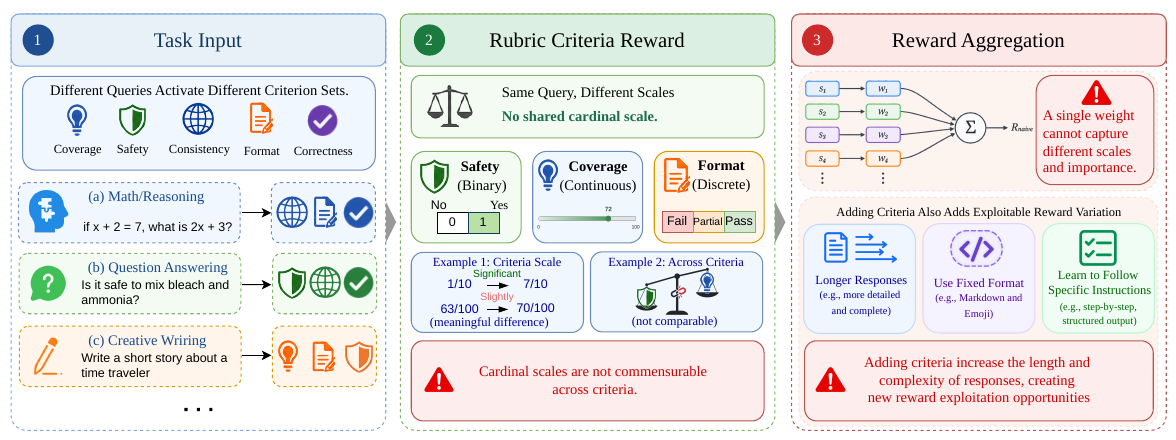}
    \caption{\textbf{Limitations of scalar reward aggregation in current RLVR pipeline.} Different queries activate different criterion sets \textbf{(left)}. The resulting binary, continuous, and discrete criterion scores have different scales, making cardinal values difficult to compare \textbf{(middle)}. Linear aggregation then conflates criterion importance with scale correction, while adding criteria can introduce exploitable reward variation such as verbosity and formatting preferences \textbf{(right)}.}
    \label{fig:background}
    \vspace{-1em} 
\end{figure*}

Early work aggregates criterion scores through a static linear combination. Uniform weighting is the most common instantiation, treating every criterion as equally decisive \citep{DBLP:conf/acl/LiuXYHYZW26, DBLP:conf/acl/GuptaSZMGBYR25, DBLP:journals/corr/abs-2602-14069}; more refined variants elicit non-uniform coefficients from domain experts or predict them with a strong model conditioned on the query and its rubric \citep{DBLP:journals/corr/abs-2507-17746, DBLP:conf/nips/ViswanathanSKCN25}. Recent work instead departs from this fixed form, reweighting criteria by shortfall, volatility, or remaining headroom \citep{DBLP:conf/acl/JiaZCLJQ26, yuan2026uncoveringmitigatingaggregationinducedreward}, normalizing criterion scores to a common scale \citep{liu2026gdpo}, or suppressing criteria that the response never licensed according to their dependencies \citep{lv2026mitigating}. This adaptivity, however, is confined to the weights: the aggregator remains a sum, and every criterion verdict is trusted as reported.  We argue that two limitations remain in this line of work (shown in Figure~\ref{fig:background}).

\textbf{\emph{(i) Prior work attends to the relative importance \emph{between} criteria while neglecting discriminative resolution \emph{within} each criterion.}}
A weight is required to serve two roles at once: to express how much a criterion matters, and to render its scores commensurable with those of the others. Prior work addresses only the former, tuning relations \emph{between} criteria while leaving resolution \emph{within} each criterion unexamined. In practice, a binary compliance rubric and a continuous coverage rubric do not express quality in the same units. Rescaling them aligns their numerical ranges but does not make their semantics comparable \citep{mandler2001compromises}. Preference-based feedback avoids the need for an absolute score scale by construction \citep{lambert2026reinforcementlearninghumanfeedback}, but it records only which response was preferred and not the criterion-specific grounds for that preference \citep{lambert2024alignmentceilingobjectivemismatch}. The evaluation standard is therefore implicit and may vary across comparisons \citep{shen2023trickledownimpactrewardinconsistency}.

\textbf{\emph{(ii) Prior work improves reward fidelity by adding criteria, but lacks a mechanism for controlling non-objective sources of reward variation.}} Reward fidelity is pursued by introducing successively more criteria, under the assumption that denser supervision yields a more faithful reward. Under linear aggregation, however, each additional criterion contributes noise alongside information. A high score on an easily satisfied criterion can offset a violation of a demanding one. For example, a response that fails to follow instructions may still receive a high reward if it scores well on formatting, organization, or length-related criteria \citep{fu2025reward}. The resulting reward surface admits exploitable optima: a policy may increase its score by producing longer responses or reproducing rewarded structural templates \citep{wang2026thinking, wang2026reward}, without improving the requested substance.


To address these limitations, we introduce \textbf{Reinforcement Learning with Verifiable Rubric-based Ranking} (\textbf{RLVR$^{2}$}), which combines the comparative nature of preference feedback with the verifiability of rubric-based rewards by representing each criterion ordinally rather than cardinally. 
\emph{To address limitation~(i)}, RLVR$^{2}$ replaces cardinal-score aggregation with within-query comparisons: each criterion induces an ordering among the rollouts of the same request, from which we estimate a latent utility on a shared scale. This removes the need for cross-criterion scale calibration, allowing the aggregation weight to reflect criterion importance rather than scale correction. \emph{To address limitation~(ii)}, RLVR$^{2}$ controls systematic reward variation from auxiliary attributes, such as reasoning length and formatting, without treating them as additional quality criteria. Specifically, these attributes enter the within-query comparison model as adjustment variables, allowing their effects on observed preferences to be estimated and removed from the quality utility. The resulting reward therefore reflects the underlying rubric quality while reducing incentives to exploit superficial attributes.


In summary, our main contributions are as follows: \textbf{\emph{(i)} We introduce RLVR$^{2}$, a verifiable ranking paradigm for rubric-based RLVR.} By replacing cardinal reward aggregation with within-group comparisons, RLVR$^{2}$ provides a principled training framework for heterogeneous rubric-based objectives, preserving criterion-level ordering while separating criterion importance from score-scale correction. \textbf{\emph{(ii)} We develop an extensible framework for reward variation control.} Heterogeneous verifiers and auxiliary attributes, such as reasoning length and formatting, are incorporated into a unified comparison model without becoming additional training objectives.  \textbf{\emph{(iii)} We extensively evaluate RLVR$^{2}$ across 3 model scales and 20 benchmarks.} It achieves leading performance on the vast majority of benchmarks consistently across model scales.

\section{Preliminaries}
\label{sec:setting}

For tasks evaluated by multiple verifiable criteria, RLVR rests on two principal components: rubric-based judgment and reward aggregation for policy optimization. We detail each below.

\paragraph{Rubric-based judgment.}
Let $\pi_\theta$ be an LLM policy parameterized by $\theta$ and a dataset $\mathcal{D}$. For each query $x\in\mathcal{D}$, the policy samples a group of $G$ candidate rollouts $\{y_i\}_{i=1}^{G}\sim\pi_\theta(\cdot\mid x)$. When a single correctness signal is insufficient, each query $x$ is evaluated using $K_x$ verifiable criteria $\mathcal{C}(x)=\{c^{(k)}\}_{k=1}^{K_x}$, where $K_x$ may vary across queries. The rubric are defined once from the query rather than resampled during training, so all $G$ rollouts of a group are graded against the same criteria. Each criterion carries a weight $w_k>0$ reflecting its importance and is scored on its own domain $\mathcal{S}_k$, which may be binary for rule-based checks, discrete for graded judgments, or continuous for metric-based measurements. We write $s_{k,i}\in\mathcal{S}_k$ for the score of rollout $y_i$ on criteria $c^{(k)}$.

\paragraph{Reward Aggregation.}
To obtain a scalar training signal, each score is mapped into a common
interval by a per-criterion map, and the standard reward is their weighted average:
\begin{equation}
R(x,y_i)=\frac{\sum_{k=1}^{K_x} w_k\,\phi_k(s_{k,i})}
{\sum_{k=1}^{K_x} w_k}\in[0,1].
\label{eq:reward}
\end{equation}
 The rubric does not specify $\{\phi_k\}$, whose choice can change $R$ even when they preserve the within-criterion orderings. Following GRPO~\citep{DBLP:journals/corr/abs-2402-03300}, the scalar reward $R_i=R(x,y_i)$ of each rollout is standardized within the group into a group-relative advantage $\hat{A}_i=(R_i-\mathrm{mean}(\bm R))/\mathrm{std}(\bm R)$, which is then used to update the policy with the following regularized clipped objective:
\begin{equation}
\begin{split}
\mathcal{J}_{\mathrm{GRPO}}(\theta)=
\mathbb{E}_{x\sim\mathcal{D},\,\{y_i\}_{i=1}^{G}\sim\pi_{\theta_{\mathrm{old}}}(\cdot\mid x)}
\Bigg[&\frac{1}{G}\sum_{i=1}^{G}\frac{1}{|y_i|}\sum_{t=1}^{|y_i|}
\min\!\big(r_{i,t}(\theta)\hat{A}_i,\\
&\quad\mathrm{clip}(r_{i,t}(\theta),1-\epsilon,1+\epsilon)\hat{A}_i\big)
-\beta\,\mathbb{D}_{\mathrm{KL}}(\pi_\theta\|\pi_{\mathrm{ref}})\Bigg].
\end{split}
\label{eq:grpo}
\end{equation}
Here $r_{i,t}(\theta)$ is the token-level importance ratio, $\epsilon$ the clipping threshold, and $\beta$ the KL weight against a reference policy $\pi_{\mathrm{ref}}$. Since $R_i$ is a rollout-level scalar, the same $\hat{A}_i$ applies to all tokens of $y_i$, so policy updates are driven entirely by $\{\phi_k\}$ and $\{w_k\}$.

\section{Methodology}
\label{sec:methodology}
We present \textbf{RLVR$^{2}$}, a verifiable ranking framework for training with heterogeneous criteria through within-group comparisons. Section~\ref{sec:ordinal-fusion} details within-group ordinal reward aggregation mechanism. Section~\ref{sec:objective-control} shows how non-objective properties fold in as covariates.
\begin{figure*}[t]
    \centering
    \includegraphics[width=\textwidth]{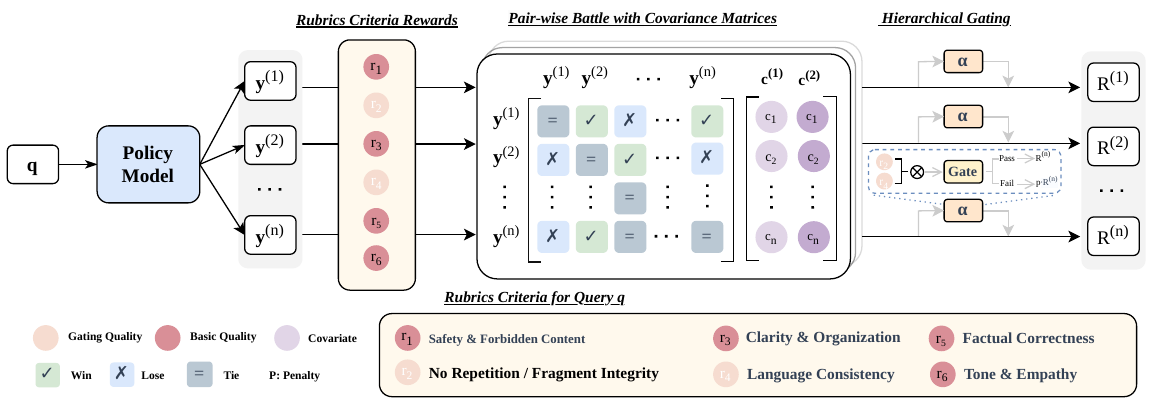}
    \caption{\textbf{Overview of RLVR$^{2}$.} Each rubric criterion becomes a within-group \emph{ordering} (\textcolor{green!60!black}{win}/\textcolor{blue}{lose}/\textcolor{gray}{tie}); orderings are fused into a scale-invariant reward without raw magnitudes. Auxiliary attributes fold in as extra \textcolor{violet}{covariates} so that hackable attributes are controlled for rather than rewarded.}
    \label{fig:methods}
\end{figure*}

\subsection{Within-Group Ordinal Reward Aggregation}
\label{sec:ordinal-fusion}
For a query $x$, let $\mathcal{Y}(x)=\{y_1,\ldots,y_G\}$ denote the rollout group sampled for policy optimization. For each of the $K_x$ verifiable criteria, the corresponding verifier assigns a score $s_{k,i}$ to rollout $y_i$. These scores may be binary, discrete, or continuous and need not share a common numerical scale. RLVR$^{2}$ uses the relative structure induced within $\mathcal{Y}(x)$: for each compared pair $(i,j)$ and criterion $k$, we encode the verifier outcome as
\begin{equation}
  o_{k}(i,j)=
  \begin{cases}
    0.5, & |s_{k,i}-s_{k,j}|\le\tau_k,\\[2pt]
    1,   & s_{k,i}-s_{k,j}>\tau_k,\\[2pt]
    0,   & s_{k,j}-s_{k,i}>\tau_k,
  \end{cases}\label{eq:battle}
\end{equation}
where $\tau_k\ge 0$ is a criterion-specific tie margin, with $\tau_k=0$ for discrete criteria. Only thresholded within-group ordering is retained, while the raw score gap is discarded. We fit a Bradley--Terry (BT) model~\citep{sun2024rethinking} to these outcomes, positing a latent utility $u_{k,i}$ per rollout with
\begin{equation}
  \Pr\!\big[\,y_i \succ y_j \mid k\,\big]
  = \sigma\!\left(u_{k,i}-u_{k,j}\right),
  \qquad
  \sigma(z)=\frac{1}{1+e^{-z}},
  \label{eq:bt}
\end{equation}
and recover $\bm{u}_k=(u_{k,1},\dots,u_{k,G})$ by maximizing the BT likelihood over the encoded outcomes. A tie outcome $o_k(i,j)=0.5$ is treated as a half-win target, encouraging the predicted probability to be one half. Since the construction uses within-group order rather than raw score magnitudes, it applies uniformly to binary, discrete, and continuous verifier outputs without cross-verifier calibration. Finally, we fuse the criterion-specific utilities and normalize them within the rollout group:
\begin{equation}
  u_i=\sum_{k=1}^{K_x}w_k u_{k,i},\qquad
  \tilde R_i=\frac{u_i-\min_j u_j}{\max_j u_j-\min_j u_j},
  \label{eq:fuse}
\end{equation}


\subsection{Objective-Preserving Attribute Adjustment}
\label{sec:objective-control}
When a new training requirement arises, a natural solution is to add another reward dimension. This may unintentionally turn an auxiliary property into an optimization target and expand the surface on which the policy can exploit the reward. RLVR$^{2}$ instead controls such attributes through the within-group comparison matrix already induced by the rubric verifiers, without expanding the rubric or changing the underlying quality objective. More generally, any measurable auxiliary attribute can be incorporated as a source of systematic variation in the observed ordering rather than introduced as a new criterion; the attributes discussed below are representative instances of this broader construction.

\paragraph{Attribute adjustment.}
For a response $y_i$, let $\bm a_i=\phi(y_i)$ denote an auxiliary attribute that is not itself part of the target rubric. We construct a normalized contrast for each attribute,
\begin{equation}
  z_{ij}^{(d)}=\frac{a_i^{(d)}-a_j^{(d)}}{a_i^{(d)}+a_j^{(d)}+\epsilon},
  \label{eq:attribute-contrast}
\end{equation}
and standardize each feature over the comparisons in the group. The resulting attributes are incorporated into the same within-group comparison matrix as the rubric-induced outcomes:
\begin{equation}
  \Pr\!\big[\,y_i\succ y_j\mid k\,\big]
  =\sigma\!\left(u_{k,i}-u_{k,j}+\bm\gamma_k^{\top}\bm z_{ij}\right).
  \label{eq:attribute-bt}
\end{equation}
Here $u_{k,i}$ represents criterion quality, whereas $\bm\gamma_k$ captures the systematic association between the auxiliary attributes and the observed ordering. Both are estimated jointly, so the attributes influence the quality estimation without becoming additional reward dimensions. 

\paragraph{Quality-level gating.}
Some requirements are non-compensatory and should not be offset by gains on other criteria. We therefore apply a stepwise adjustment after ordinal fusion. Let $\tilde R_i$ denote the normalized fused reward for rollout $i$. A failure of the global-quality condition triggers a hard gate, while the low-quality condition yields a soft penalty factor $\rho_i$:
\begin{equation}
    R_i = g(q_i)\tilde R_i,
    \qquad
    g(q_i)=
  \begin{cases}
    0, & \min(\bm s_i^{\mathrm{global}})<1,\\[2pt]
    \rho_i\tilde R_i, & \text{otherwise},
  \end{cases}
  \qquad 0<\rho_i\leq 1,
  \label{eq:quality-gate}
\end{equation}
where $\bm s_i^{\mathrm{global}}$ contains the GlobalLowQuality scores. For the LowQuality layer, let $\bar s_i^{\mathrm{low}}$ be its mean score. We set
\begin{equation}
  \rho_i=
  \begin{cases}
    1, & \bar s_i^{\mathrm{low}}\geq\tau,\\[2pt]
    \rho_{\min}+(1-\rho_{\min})\,\operatorname{softgate}(\bar s_i^{\mathrm{low}}),
    & \bar s_i^{\mathrm{low}}<\tau,
  \end{cases}
  \label{eq:quality-penalty}
\end{equation}
where $\tau$ is the low-quality threshold and $\rho_{\min}$ is the penalty floor.

\paragraph{Reasoning-efficiency control.}
The adjustment is useful when rubrics reward information coverage, as in recall-oriented and question-answering tasks. Once the relevant content has been identified, additional deliberation may yield repeated verification or marginal rubric coverage without improving substantive quality. Accordingly, a positive association between reasoning length and observed preference need not reflect a genuine quality benefit. We randomly attenuate a fraction $\lambda$ of the fitted positive coefficient, while leaving non-positive coefficients unchanged. This controls the strength of adjustment without imposing a target response length, thereby reducing unnecessary deliberation while limiting abrupt changes in reasoning behavior.

\paragraph{Formatting control.}                                   
Judges may systematically prefer heavily structured responses, even when extra formatting adds no substantive information~\citep{yu2025dapoopensourcellmreinforcement}. We therefore treat formatting as an auxiliary source of variation, not a quality criterion. Normalized statistics for headings, lists, and emphasis are standardized and included in the attribute-adjusted model, controlling formatting preferences without directly rewarding formatting.
\section{Experiments}

\label{sec:experiments}
In this section, we present the empirical evaluation of \textbf{RLVR$^{2}$}. We first describe the experimental setup (Section~\ref{sec:setup}), then report the main results (Section~\ref{sec:main}), further analyze attribute adjustment (Section~\ref{sec:further-analysis}), and conclude with ablations of its robustness and computational cost (Section~\ref{sec:ablation}).

    
    
    
    


\subsection{Experimental Setup}
\label{sec:setup}
\paragraph{Datasets and Benchmarks.}
All methods are trained on the same 120k-query mixture of general-chat, scientific reasoning and agentic tasks. They share rubrics, verifiers, and judges and differ only in reward aggregation. For evaluation, we use a broad suite of public benchmarks covering six capability dimensions: \emph{instruction following} (SysBench [SysB] \citep{qin2024sysbench}, AdvancedIF [AdvIF] \citep{he2026advancedif}, IFEval [IFE] \citep{zhou2023instruction}, IFBench [IFB] \citep{pyatkin2026generalizing}, and MultiChallenge [MChal] \citep{deshpande2025multichallenge}), \emph{long-context} (LongBench-v2 [LB-v2] \citep{bai2025longbench}, LongProc [LProc] \citep{ye2025longproc}, HELMET [HLM] \citep{yen2025helmet}, Frames [Frm] \citep{krishna2024factfetchreasonunified}, AA-LCR [AA-L] \citep{artificialanalysis2025lcr}), \emph{factuality and QA} (SimpleQA-verified [SQA-v] \citep{haas2025simpleqa}, ChineseSimpleQA [CSQA] \citep{he2025chinese}), \emph{reasoning} (GPQA \citep{rein2023gpqa}), \emph{knowledge} (MMLU-Pro [MMLU-P] \citep{wang2024mmlu}), and \emph{open-ended chat} (ArenaHard-v2 [AH-v2] \citep{li2024crowdsourced}, EQBench3 [EQB3] \citep{paech2023eqbench}) and \emph{instruction following}  ($\tau^3$-Banking \citep{shi2026tau}, GAIA \citep{mialon2024gaia}, GDPVal \citep{patwardhan2026gdpval}, SWE-bench Verified [SWE-V]\citep{jimenez2024swebench}). Dataset and benchmarks are detailed in Appendix~\ref{appendix:dataset}.

\paragraph{Baselines and Implementation Details.} Beyond the vanilla model, we compare RLVR$^{2}$ with several static and adaptive reward-fusion baselines: Rule-Fusion~\citep{park2024max}, a weighted sum gated by hard rules; the graph-based gating method GEAR~\citep{lv2026mitigating}; and two dynamic aggregation methods, Focal~\citep{DBLP:journals/corr/abs-2508-12790} and GDPO~\citep{liu2026gdpo} \footnote{AMRP \citep{yuan2026uncoveringmitigatingaggregationinducedreward} is excluded, as no official implementation was publicly available.}. We evaluate our method on a diverse set of LLMs spanning two model families and different scales, including the Qwen3.5 series (Qwen3.5-4B and Qwen3.5-9B) \citep{qwen3.5} and GLM-4.5-Air (106B-A12B) \citep{zeng2025glm}. The models cover both dense and mixture-of-experts(MoE) architectures. All methods are implemented using the PaddleRL framework and trained with the GSPO algorithm \citep{zheng2025groupsequencepolicyoptimization} on up to $128\times$ H800-80GB GPUs. Unless otherwise specified, we use a global batch size of 128, a context length of 256K, and a group size of $G=8$ rollouts per query. We optimize all models with AdamW using a learning rate of $1\times10^{-6}$. RLVR$^{2}$-specific settings are held fixed across models and listed in Appendix~\ref{appendix:train_dataset}.

\subsection{Main Results}
\label{sec:main}
\begin{table*}[t]
  \centering
  \tiny
  \setlength{\tabcolsep}{2.4pt}
  \renewcommand{\arraystretch}{1.15}
  \resizebox{\textwidth}{!}{
  \begin{tabular}{clcccccccccccccccccc}
    \toprule
    & \multirow[c]{2}{*}[-0.7ex]{\textbf{Method}}
      & \multicolumn{5}{c}{\textbf{Instruction Following}}
      & \multicolumn{2}{c}{\textbf{Factuality \& QA}}
      & \multicolumn{4}{c}{\textbf{Long-Context}}
      & \multicolumn{2}{c}{\textbf{Reason.}}
      & \multicolumn{1}{c}{\textbf{Know.}}
      & \multicolumn{2}{c}{\textbf{Open-ended}}
      & \multirow[c]{2}{*}[-0.7ex]{\textbf{Average}} \\
    \cmidrule(lr){3-7}\cmidrule(lr){8-9}\cmidrule(lr){10-13}\cmidrule(lr){14-15}\cmidrule(lr){16-16}\cmidrule(lr){17-18}
    &
      & SysB & AdvIF & IFE & IFB & MChal
      & SQA-v & CSQA
      & LB-v2 & LProc & HLM & Frm
      & AA-L & GPQA
      & MMLU & AH-v2 & EQB3 & \\
    \midrule
    & Base        & $51.08$ & $43.93$ & $76.16$ & $36.05$ & $40.83$ & $12.6$ & $51.37$ & $54.59$ & $81.08$ & $29.74$ & $77.43$ & $47$ & $71.46$ & $\underline{79.19}$ & $52.6$ & $85.35$ & $55.65$ \\
    & Rule-Fusion & $66.88$ & $55.11$ & $82.42$ & $67.01$ & $52.64$ & $13.9$ & $53.67$ & $64.62$ & $83.55$ & $31.38$ & $78.37$ & $63.33$ & $74.21$ & $78.73$ & $61.61$ & $82.62$ & $63.13$ \\
    & GEAR        & $41.6$ & $58.61$ & $93.9$ & $62.93$ & $\underline{59.17}$ & $6.6$ & $52.62$ & $63.94$ & $\underline{85.27}$ & $32.47$ & $\underline{80.46}$ & $62.33$ & $76.26$ & $77.41$ & $65.2$ & $85.3$ & $62.75$ \\
    & GDPO        & $71.44$ & $62.18$ & $\underline{94.27}$ & $71.77$ & $50.14$ & $9.1$ & $53.21$ & $57.38$ & $83.45$ & $32.94$ & $80.1$ & $62.33$ & $75$ & $75.36$ & $62.9$ & $82.65$ & $64.01$ \\
    & GDPO+       & $70$ & $62.92$ & $91.68$ & $\mathbf{73.47}$ & $57.5$ & $9.7$ & $54.17$ & $57.97$ & $83.06$ & $31.88$ & $79.85$ & $\underline{65}$ & $75.63$ & $76.79$ & $\mathbf{67.6}$ & $83.6$ & $65.05$ \\
    & Focal       & $68.6$ & $59.89$ & $92.24$ & $72.79$ & $48.95$ & $13.8$ & $51.47$ & $\mathbf{67.69}$ & $81.04$ & $33.37$ & $79.25$ & $63$ & $75.48$ & $75.44$ & $65.98$ & $84.13$ & $64.57$ \\
    & Focal+      & $\underline{72.52}$ & $\underline{65.7}$ & $91.5$ & $65.99$ & $58.67$ & $\underline{14}$ & $\underline{55.23}$ & $64.62$ & $\mathbf{85.28}$ & $32.42$ & $\underline{80.46}$ & $59$ & $76.61$ & $78.65$ & $66.65$ & $84.32$ & $\underline{65.73}$ \\
    \rowcolor{scaleBlue!13}
    & RLVR$^{2}$        & $69.12$ & $64.64$ & $93.16$ & $70.71$ & $53.19$ & $13.8$ & $53.97$ & $63.85$ & $84.02$ & $\underline{34.36}$ & $\mathbf{81.19}$ & $64$ & $\underline{77.2}$ & $73.55$ & $66.74$ & $\underline{86.04}$ & $65.60$ \\
    \rowcolor{scaleTeal!16}
    \multirow{-9}{*}{\emph{106B}} & RLVR$^{2}$+ & $\mathbf{73.2}$ & $\mathbf{67.32}$ & $\mathbf{94.64}$ & $\underline{73.13}$ & $\mathbf{60.04}$ & $\mathbf{15.2}$ & $\mathbf{56.07}$ & $\underline{65.77}$ & $84.12$ & $\mathbf{35.31}$ & $\mathbf{81.19}$ & $\mathbf{67.67}$ & $\mathbf{77.86}$ & $\mathbf{80.43}$ & $\underline{67.14}$ & $\mathbf{87.42}$ & $\mathbf{67.91}$ \\
    \midrule
    & Base        & $67.44$ & $59.02$ & $89.65$ & $42.77$ & $50$ & $11.5$ & $67.23$ & $32.8$ & $80.98$ & $24.87$ & $74.76$ & $54.33$ & $80.18$ & $80.33$ & $35.1$ & $82.75$ & $58.36$ \\
    & Rule-Fusion & $74.12$ & $62.23$ & $90.76$ & $48.59$ & $\underline{55.56}$ & $11.6$ & $67.13$ & $32.01$ & $81.75$ & $25.86$ & $74.64$ & $55.33$ & $80.68$ & $80.39$ & $60.1$ & $86.65$ & $61.71$ \\
    & GEAR        & $73.68$ & $61.55$ & $\underline{92.05}$ & $44.28$ & $50$ & $\underline{12.1}$ & $67.6$ & $\underline{33.6}$ & $80.69$ & $4.17$ & $\underline{76.58}$ & $58.33$ & $82.45$ & $80.63$ & $70.3$ & $\mathbf{88}$ & $61.00$ \\
    & GDPO        & $74.88$ & $\underline{63.71}$ & $89.83$ & $67.35$ & $52.37$ & $11.2$ & $68.17$ & $31.61$ & $81.53$ & $27.62$ & $75.97$ & $57.33$ & $\mathbf{83.33}$ & $\mathbf{81.68}$ & $62.3$ & $82.55$ & $63.21$ \\
    & GDPO+       & $68.76$ & $51.34$ & $72.46$ & $59.52$ & $50$ & $10.1$ & $54.27$ & $31.41$ & $\underline{82.55}$ & $25.42$ & $71.97$ & $50.33$ & $82.07$ & $80.15$ & $67.3$ & $84.55$ & $58.89$ \\
    & Focal       & $\mathbf{76.44}$ & $63.3$ & $91.31$ & $69.05$ & $51.67$ & $10.8$ & $68.3$ & $\underline{33.6}$ & $80.93$ & $\underline{27.76}$ & $76.46$ & $57$ & $\underline{82.7}$ & $81.01$ & $61$ & $86.2$ & $63.60$ \\
    & Focal+      & $\underline{75.2}$ & $61.19$ & $90.57$ & $\underline{70.41}$ & $53$ & $10.9$ & $65.53$ & $31.81$ & $81.91$ & $26.76$ & $\mathbf{76.94}$ & $\underline{59.67}$ & $81.44$ & $80.73$ & $57.3$ & $86.6$ & $63.12$ \\
    \rowcolor{scaleBlue!13}
    & RLVR$^{2}$        & $73.48$ & $63.61$ & $\underline{92.05}$ & $69.69$ & $55.40$ & $11.1$ & $\underline{68.43}$ & $\mathbf{33.8}$ & $81.75$ & $\mathbf{28.18}$ & $75.73$ & $58.67$ & $80.81$ & $80.48$ & $\underline{71.1}$ & $\underline{87.15}$ & $\underline{64.46}$ \\
    \rowcolor{scaleTeal!16}
    \multirow{-9}{*}{\emph{9B}} & RLVR$^{2}$+ & $\mathbf{76.44}$ & $\mathbf{65.56}$ & $\mathbf{92.61}$ & $\mathbf{73.81}$ & $\mathbf{56.23}$ & $\mathbf{12.2}$ & $\mathbf{69.5}$ & $32.21$ & $\mathbf{82.77}$ & $26$ & $76.21$ & $\mathbf{60}$ & $81.94$ & $\underline{81.11}$ & $\mathbf{72.7}$ & $85.65$ & $\mathbf{65.31}$ \\
    \midrule
    & Base        & $61$ & $46.51$ & $83.55$ & $58.84$ & $33.75$ & $6.9$ & $\underline{56.67}$ & $\underline{32.31}$ & $73.22$ & $23.5$ & $73.18$ & $54.33$ & $77.65$ & $76.53$ & $28.1$ & $77.3$ & $53.96$ \\
    & Rule-Fusion & $57.8$ & $58.36$ & $91.13$ & $68.71$ & $50$ & $7.2$ & $55.53$ & $32.01$ & $71.74$ & $23.7$ & $72.33$ & $55$ & $76.77$ & $76.74$ & $\underline{48.5}$ & $82.35$ & $57.99$ \\
    & GEAR        & $\underline{72.12}$ & $59.48$ & $\underline{91.68}$ & $46.08$ & $33.33$ & $5$ & $54.8$ & $\mathbf{32.6}$ & $\underline{80.45}$ & $23.44$ & $73.67$ & $54$ & $78.28$ & $77.29$ & $41.3$ & $\mathbf{85.5}$ & $56.81$ \\
    & GDPO        & $52.48$ & $52.62$ & $83.73$ & $64.29$ & $\underline{56.16}$ & $6.9$ & $54$ & $31.41$ & $79.71$ & $23.51$ & $70.51$ & $46.67$ & $79.01$ & $75.21$ & $45.8$ & $83.65$ & $56.60$ \\
    & GDPO+       & $42.76$ & $46.64$ & $74.12$ & $56.8$ & $35.42$ & $6.2$ & $55.23$ & $30.82$ & $77.1$ & $24.34$ & $73.42$ & $52.33$ & $\underline{79.03}$ & $76.66$ & $43.4$ & $83.6$ & $53.62$ \\
    & Focal       & $67.28$ & $55.58$ & $\underline{91.68}$ & $66.33$ & $53.06$ & $\underline{7.8}$ & $56.43$ & $29.82$ & $\mathbf{81.79}$ & $\underline{26.47}$ & $\underline{74.39}$ & $45.67$ & $77.78$ & $77.07$ & $44.5$ & $84.15$ & $58.74$ \\
    & Focal+      & $70.24$ & $59.83$ & $91.5$ & $\underline{70.83}$ & $53.22$ & $7.3$ & $56.47$ & $32.01$ & $78.43$ & $26.26$ & $74.03$ & $\underline{55.67}$ & $77.02$ & $\underline{77.63}$ & $41.3$ & $85.15$ & $59.81$ \\
    \rowcolor{scaleBlue!13}
    & RLVR$^{2}$        & $67.12$ & $\underline{61.09}$ & $90.76$ & $\mathbf{71.77}$ & $54$ & $7.2$ & $56.66$ & $31.21$ & $76.57$ & $25.31$ & $\mathbf{74.51}$ & $54.67$ & $77.15$ & $76.8$ & $\mathbf{50.8}$ & $\underline{85.2}$ & $\underline{60.05}$ \\
    \rowcolor{scaleTeal!16}
    \multirow{-9}{*}{\emph{4B}} & RLVR$^{2}$+ & $\mathbf{74.4}$ & $\mathbf{62.6}$ & $\mathbf{92.42}$ & $69.73$ & $\mathbf{57.29}$ & $\mathbf{8.2}$ & $\mathbf{56.86}$ & $32.21$ & $78.61$ & $\mathbf{26.71}$ & $\underline{74.39}$ & $\mathbf{60.33}$ & $\mathbf{79.04}$ & $\mathbf{77.67}$ & $46.5$ & $83.5$ & $\mathbf{61.28}$ \\
    \bottomrule
  \end{tabular}
  }
  \caption{\textbf{Comparison of reward-aggregation methods across 16 benchmarks and three model scales.} Within each block, all methods share the same training setting, differing only in reward aggregation. \textbf{Bold} and \underline{underline} mark the best and second-best per block.}
  \label{tab:main}
\end{table*}

\paragraph{Performance across capabilities and scales.}
As shown in Table~\ref{tab:main}, RLVR$^{2}$ performs consistently across all three model scales, leading a majority of benchmarks and showing particular strength in instruction following and factuality, alongside broad gains in long-context understanding, reasoning, knowledge, and open-ended evaluation. This pattern reflects the benefit of within-group ordinal fusion on tasks governed by multiple criteria: unlike cardinal aggregation, it prevents a high score on one dimension from masking a critical failure on another. Consequently, responses that are fluent but factually incorrect or instruction-violating are less likely to receive inflated rewards.

\begin{table}[htbp]
  \centering
  \label{tab:composable}
  \begin{subtable}[t]{0.49\linewidth}
    \centering
    \scriptsize
    \setlength{\tabcolsep}{2pt}
    \renewcommand{\arraystretch}{1.15}
    \caption{Agentic and coding performance.}
    \resizebox{\linewidth}{!}{%
    \begin{tabular}{@{}lccccc@{}}
      \toprule
      \textbf{Method} & \textbf{$\tau^3$-Banking} & \textbf{GAIA} & \textbf{GDPVal} & \textbf{SWE-V} \\
      \midrule
      RLVR$^{2}$  & $7.53$ & $62.14$ & $56.86$ & $52.62$ \\
      Focal       & $7.12$ & $59.37$ & $54.95$ & $50.88$ \\
      GDPO        & $6.95$ & $55.84$ & $54.20$ & $48.67$ \\
      Rule-Fusion & $7.22$ & $58.25$ & $52.96$ & $48.69$\\
      Base        & $6.91$ & $53.37$ & $51.38$ & $49.40$\\
      \bottomrule
    \end{tabular}%
    }
    \label{tab:focal}
  \end{subtable}\hfill
  \begin{subtable}[t]{0.49\linewidth}
    \centering
    \scriptsize
    \setlength{\tabcolsep}{1pt}
    \renewcommand{\arraystretch}{1.15}
    \caption{Reasoning-efficiency control.}
    \resizebox{0.96\linewidth}{!}{%
    \begin{tabular}{@{}llccccccc@{}}
      \toprule
      \textbf{Method} & \textbf{Setting} & \textbf{IF} & \textbf{Fa.} & \textbf{Lo.}
        & \textbf{Re.} & \textbf{Kn.} & \textbf{Op.} & \textbf{Ovr.} \\
      \midrule
      Rule-Fusion & --
        & $68.93$ & $39.37$ & $53.57$ & $68.01$ & $80.39$ & $86.65$ & $66.15$ \\
      \cmidrule(l){1-9}
      \multirow{6}{*}{RLVR$^{2}$+} & --
        & $77.11$ & $40.85$ & $54.24$ & $70.97$ & $81.11$ & $85.65$ & $68.32$ \\
      & $\lambda{=}0$
        & $75.73$ & $40.37$ & $53.73$ & $71.01$ & $80.85$ & $85.85$ & $67.92$ \\
      & $\lambda{=}0.25$
        & $75.28$ & $42.37$ & $54.07$ & $70.21$ & $80.72$ & $87.30$ & $68.30$ \\
      & $\lambda{=}0.50$
        & $75.78$ & $41.75$ & $53.56$ & $70.10$ & $81.41$ & $85.30$ & $67.98$ \\
      & $\lambda{=}0.75$
        & $77.73$ & $41.86$ & $54.42$ & $70.49$ & $81.31$ & $85.60$ & $68.57$ \\
     \midrule
      & $Format$
        & $76.32$ & $41.62$ & $53.72$ & $70.07$ & $80.21$ & $85.76$ & $67.95$ \\
      \bottomrule
    \end{tabular}%
    }
    \label{tab:appendix_think_compr_avg}
  \end{subtable}
  \caption{\textbf{Further analysis of RLVR$^{2}$.} Bold denotes the best value in each column.}
\end{table}

\paragraph{Reasoning performance.}
Reasoning tasks evaluate more than the final answer: they require understanding the problem, decomposing it into intermediate steps, maintaining consistency across a long solution process, and reaching a verifiable conclusion. The corresponding verifier signals therefore reflect several aspects of reasoning quality, including intermediate validity, mathematical or scientific correctness, and final-answer agreement. RLVR$^{2}$+ leads both AA-LCR and GPQA , while remaining highly competitive at 9B, showing that ordinal fusion effectively combines process- and outcome-level signals.

\paragraph{Agentic performance.}
We further evaluate RLVR$^{2}$ on agentic benchmarks, where verifiers assess complete multi-step trajectories rather than isolated responses. 
As shown in Table~\ref{tab:focal}, RLVR$^{2}$ achieves the best result on all four benchmarks. Its gains over the strongest competing aggregation method are $0.31$, $2.77$, $1.91$, and $1.74$ points, respectively. These consistent improvements show that within-group ordinal fusion transfers effectively to trajectory-level policy optimization, jointly capturing tool behavior, intermediate reasoning, task completion, and final response quality.

\subsection{Further Analysis}
\label{sec:further-analysis}

\subsubsection{Quality-level gating}
\paragraph{Setup.} We apply the same gate mechanism to RLVR$^{2}$ and the comparable baselines, denoting gated variants by a ``$+$'' suffix in Table~\ref{tab:main}. The gate acts on critical rubric levels after criterion fusion. 

\paragraph{Result.} The gated variant RLVR$^{2}$+ outperforms RLVR$^{2}$ in most of benchmarks comparisons across three scales
Cardinal aggregation baselines show less consistent changes. This result indicates that hard quality constraints are compatible with within-group ordinal rewards.

\subsubsection{Reasoning-efficiency control}

\begin{figure*}[htbp]
    \centering
    \includegraphics[width=\textwidth]{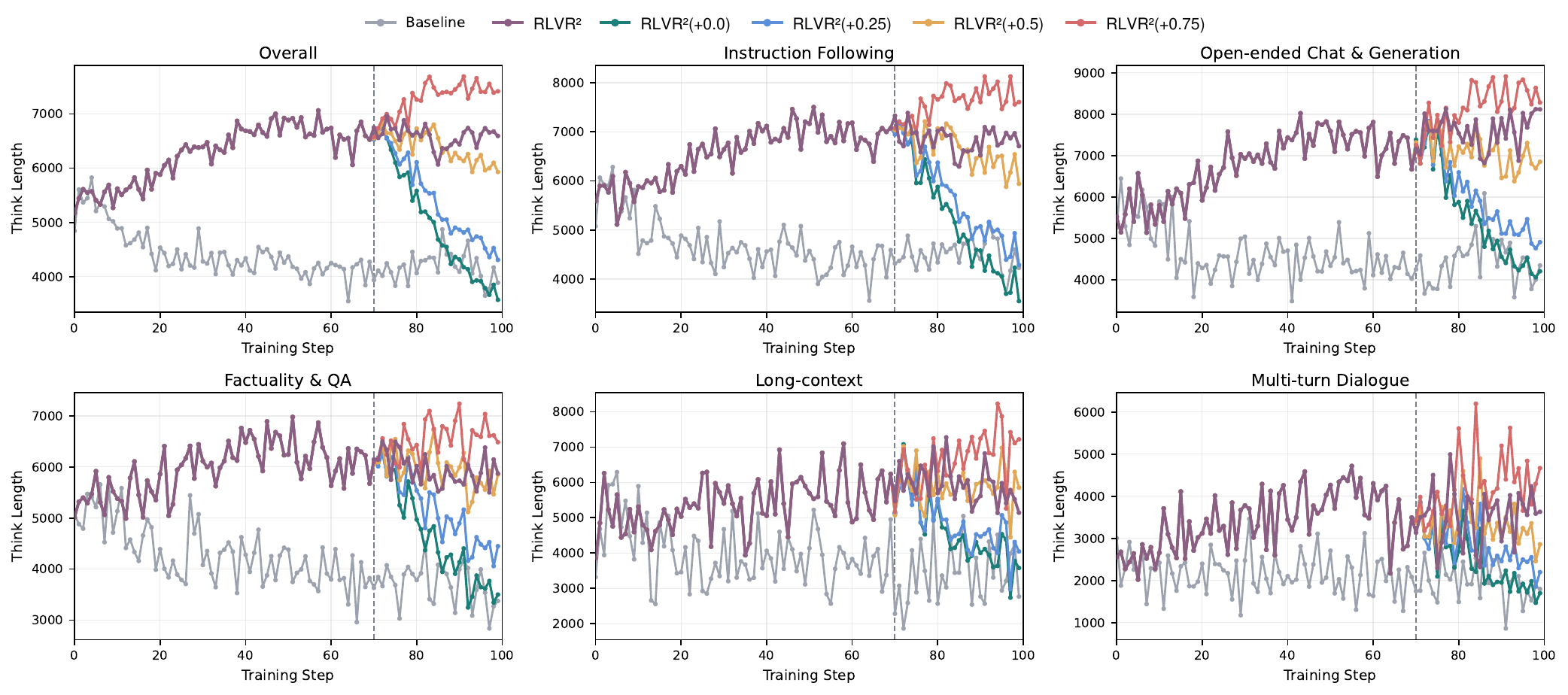}
    \caption{\textbf{Reasoning-efficiency control under different attenuation ratios.} Larger $\lambda$ yields faster convergence toward baseline length, with sensitivity varying by domain.}
    \label{fig:think_length}
\end{figure*}

\paragraph{Setup.} We resume the Qwen3.5-9B run from step~70 and vary the attenuation ratio $\lambda\in\{0,0.25,0.50,0.75\}$, while keeping the rubric criteria and optimization settings fixed. The adjustment is applied to the fitted positive association between reasoning extent and the observed rubric ordering. Table~\ref{tab:appendix_think_compr_avg} reports capability-level averages, while Figure~\ref{fig:think_length} reports thinking-token dynamics.

\paragraph{Result.} Across the tested values of $\lambda$, the adjustment changes reasoning extent while keeping the capability-level averages within a narrow range. This suggests that coefficient attenuation controls reasoning efficiency without introducing a separate reasoning-length criterion. 

\subsubsection{Formatting control}

\begin{table*}[htbp]
  \centering
  \tiny
  \setlength{\tabcolsep}{2.8pt}
  \renewcommand{\arraystretch}{1.15}
  \resizebox{\textwidth}{!}{%
  \begin{tabular}{c*{11}{c}}
    \toprule
    \textbf{Step}
      & \textbf{Heading}\,\textcolor{green!50!black}{$\downarrow$}
      & \textbf{Bold}\,\textcolor{green!50!black}{$\downarrow$}
      & \textbf{Italic}
      & \textbf{Table Row}\,\textcolor{green!50!black}{$\downarrow$}
      & \textbf{Code Block}\,\textcolor{green!50!black}{$\downarrow$}
      & \textbf{Inline Code}
      & \textbf{OL}\,\textcolor{green!50!black}{$\downarrow$}
      & \textbf{UL}\,\textcolor{green!50!black}{$\downarrow$}
      & \textbf{Blockquote}
      & \textbf{HR}\,\textcolor{green!50!black}{$\downarrow$}
      & \textbf{Link} \\
    \midrule

    \rowcolor{scaleBlue!13}
    $0$   & $34.45$ & $48.67$ & $12.79$ & $13.48$ & $7.26$ & $2.25$
      & $13.31$ & $28.80$ & $12.16$ & $26.65$ & $0.20$ \\
    \rowcolor{scaleBlue!13}
    $10$  & $34.52$ & $49.38$ & $13.15$ & $14.24$ & $7.83$ & $2.06$ & $13.54$ & $31.19$ & $12.48$ & $26.87$ & $0.45$  \\
    \rowcolor{scaleBlue!13}
    $20$  & $34.63$ & $49.79$ & $13.33$ & $14.66$ & $8.42$ & $2.01$ & $14.05$ & $32.31$ & $12.64$ & $26.27$ & $0.02$ \\
    \rowcolor{scaleBlue!13}
    $30$  & $34.71$ & $50.55$ & $14.06$ & $14.94$ & $8.54$ & $1.83$ & $14.47$ & $32.64$ & $12.71$ & $26.94$ & $0.11$ \\
    \rowcolor{scaleBlue!13}
    $40$  & $34.82$ & $50.81$ & $13.28$ & $15.13$ & $7.95$ & $2.44$ & $14.72$ & $33.01$ & $11.95$ & $27.54$ & $0.04$ \\
    \rowcolor{scaleBlue!13}
    $50$  & $34.94$ & $51.23$ & $13.44$ & $15.93$ & $9.06$ & $3.33$ & $15.20$ & $33.34$ & $12.52$ & $27.20$ & $0.31$ \\
    \rowcolor{scaleBlue!13}
    $60$  & $35.08$ & $51.73$ & $13.04$ & $16.34$ & $9.29$ & $2.17$ & $15.87$ & $34.08$ & $12.88$ & $28.03$ & $0.28$ \\
    \rowcolor{scaleBlue!13}
    $70$  & $35.21$ & $52.10$ & $13.32$ & $16.72$ & $10.76$ & $2.25$ & $16.42$ & $34.98$ & $12.24$ & $29.08$ & $0.29$ \\

    \rowcolor{scaleTeal!16}
    $80$  & $34.58$ & $50.71$ & $13.17$ & $15.65$ & $8.28$ & $2.42$ & $15.85$ & $33.36$ & $12.56$ & $27.22$ & $0.33$ \\
    \rowcolor{scaleTeal!16}
    $90$  & $34.12$ & $49.86$ & $13.20$ & $14.28$ & $6.41$ & $2.26$ & $15.40$ & $32.03$ & $12.86$ & $26.18$ & $0.18$ \\
    \rowcolor{scaleTeal!16}
    $100$ & $33.51$ & $48.92$ & $13.14$ & $13.38$ & $5.25$ & $2.21$
      & $14.22$ & $31.51$ & $12.19$ & $25.81$ & $0.24$ \\

    \rowcolor{scaleBlue!32}
    \multicolumn{1}{l}{\textbf{Avg.}}
      & $34.80$ & $50.53$ & $13.30$ & $15.18$ & $8.64$ & $2.29$
      & $14.70$ & $32.54$ & $12.45$ & $27.32$ & $0.21$ \\
    \rowcolor{scaleTeal!32}
    \multicolumn{1}{l}{\textbf{Avg.}}
      & $34.07$
      & $49.83$
      & $13.17$
      & $14.44$
      & $6.65$
      & $2.30$
      & $15.16$
      & $32.30$
      & $12.54$
      & $26.40$
      & $0.25$ \\
    \bottomrule
  \end{tabular}
  }
  \caption{\textbf{Markdown usage during RL training.} Columns report Markdown-marker density (\%); OL, UL, and HR denote ordered lists, unordered lists, and horizontal rules. Light rows show the pre- and post-control phases; dark rows show averages. Green arrows indicate post-control decreases.}
  \label{tab:style-drift}
\end{table*}

\paragraph{Setup.} We analyze the Qwen3.5-9B training trajectory, tracking densities of markdown format. The formatting-control strategy is applied after the $70$\% checkpoint by treating these statistics as auxiliary attributes in the ordinal comparison model rather than as additional rubric criteria.


\paragraph{Result.} Formatting usage rises before control but declines after attribute adjustment, while query-dependent markers remain relatively stable. The formatting-controlled variant achieves an overall capability average of $68.32$, close to the $67.95$ of RLVR$^{2}+$ in Table~\ref{tab:appendix_think_compr_avg}. These results show that formatting-related preference can be controlled without expanding the rubric or directly rewarding formatting. Detailed statistics are reported in Table~\ref{tab:style-drift}.

\subsection{Ablation Study}
\label{sec:ablation}

\paragraph{Effective weight and criterion headroom.}
\begin{figure*}[htbp]
    \centering
    \includegraphics[width=\textwidth]{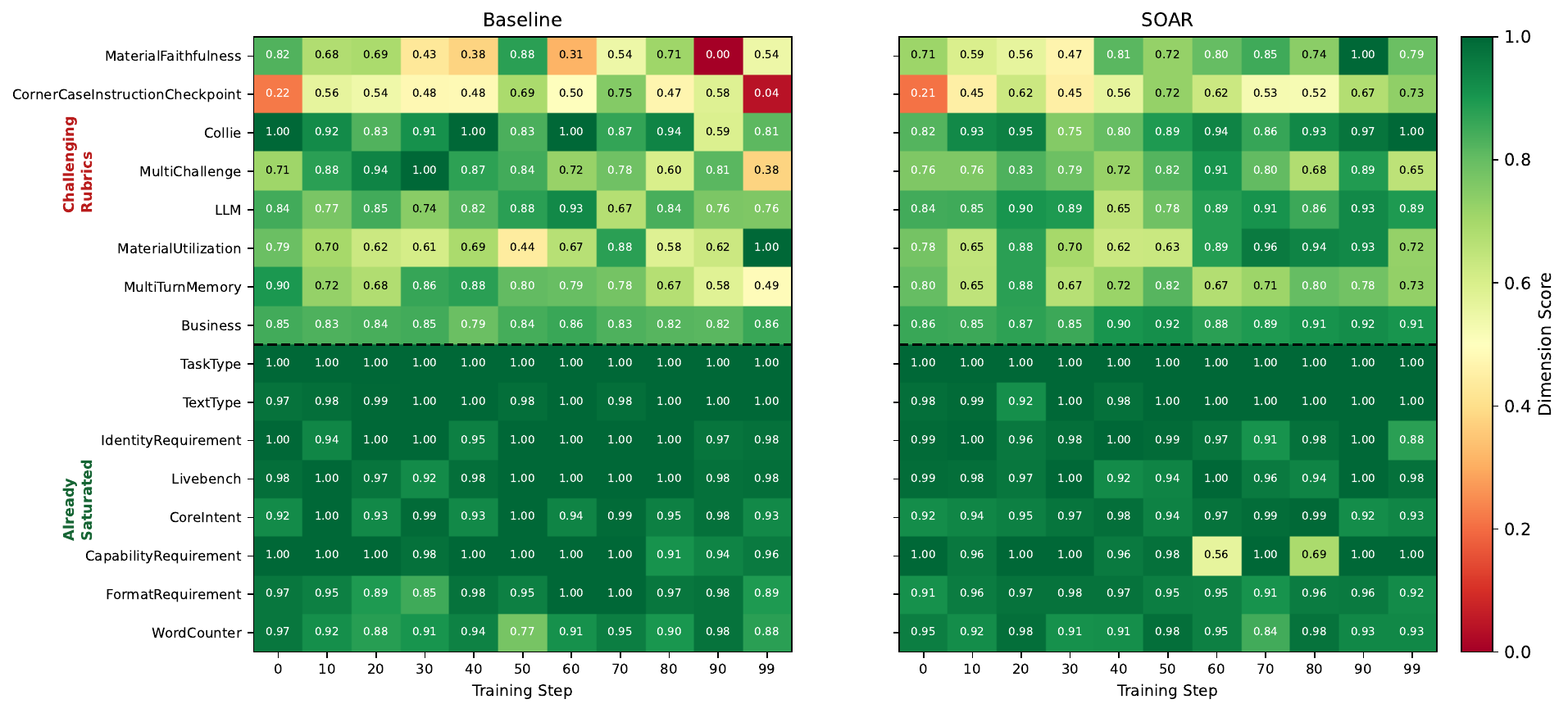}
    \caption{\textbf{Criterion-level training dynamics.} RLVR$^{2}$ improves more on Challenging Rubrics while matching Baseline on Already-Saturated ones.}
    \label{fig:headroom}
\end{figure*}

Figure~\ref{fig:headroom} compares training dynamics on challenging and already-saturated rubric subsets. RLVR$^{2}$ improves more on criteria with remaining headroom while preserving already-saturated performance. A criterion tied across an entire rollout group contributes no utility difference, so fusion emphasizes criteria that distinguish sampled responses. On the challenging subset, the baseline drops by $15.6$ points from start to end 
, whereas RLVR$^{2}$ gains $8.0$ points from a lower starting point; the saturated subset remains essentially flat for both methods.

\begin{table}[htbp]
\centering
\begin{minipage}[t]{0.5\textwidth}
\centering
\footnotesize
\setlength{\tabcolsep}{3.5pt}
\begin{tabular}{lccccccc}
\toprule
$G$ & IF & Fact. & Long & Reas. & Know. & Open & Avg. \\
\midrule
$4$  & $73.59$ & $38.77$ & $54.33$ & $70.25$ & $80.67$ & $67.68$ & $63.71$ \\
$8$  & $73.45$ & $40.04$ & $53.50$ & $71.43$ & $81.28$ & $67.80$ & $63.82$ \\
$16$ & $73.47$ & $39.04$ & $53.43$ & $68.40$ & $80.98$ & $68.33$ & $63.35$ \\
\bottomrule
\end{tabular}
\caption{\textbf{Effect of rollout group size.} Columns are dimension-level averages. }
\label{tab:group_size}
\end{minipage}
\hfill
\begin{minipage}[t]{0.45\textwidth}
\centering
\footnotesize
\setlength{\tabcolsep}{3.5pt}
\begin{tabular}{lcccc}
\toprule
Method & Len. & Fusion & Rollout & Total \\
\midrule
Rule-Fusion & $4936$ & $12.7$ & $101.0$ & $146.7$ \\
Focal       & $5570$ & $20.5$ & $149.4$ & $222.3$ \\
GDPO        & $8581$ & $16.5$ & $174.3$ & $259.9$ \\
RLVR$^{2}$     & $7344$ & $10.9$ & $160.2$ & $235.4$ \\
\bottomrule
\end{tabular}
\caption{\textbf{Runtime overhead.} Rubric inference is shared across methods and excluded.}
\label{tab:cost}
\end{minipage}
\end{table}

\paragraph{Rollout group size.}
We vary $G\in\{4,8,16\}$ for 50 training steps with all other settings fixed. Larger groups provide more within-group comparisons but incur proportional rollout cost. Table~\ref{tab:group_size} shows that RLVR$^{2}$+ is relatively stable across the tested group sizes: the overall average varies only from $63.35$ to $63.82$, with $G=8$ providing a practical trade-off between comparison density, query diversity, and computation. Per-benchmark scores are reported in Appendix~\ref{app:gropu_size_result}.

\paragraph{Computational overhead.}
Table~\ref{tab:cost} compares reward-aggregation overhead on Qwen3.5-9B with batch size 128 and $G=8$. Rubric scoring is shared and excluded; fusion is measured on identical logged rollouts, while total step time also includes generation length. RLVR$^{2}$ operates only on computed rubric outcomes and makes no additional rubric calls. Its fusion takes $10.9$ seconds per step, compared with others approaches.


\section{Related Work}

\paragraph{Rubric-based rewards for RLVR.}
Rubric-based grading has extended RLVR from tasks with well-defined correctness signals, such as mathematics and code, to open-ended generation with multiple quality requirements~\citep{yu2026dapo}. Existing studies investigate rubric construction, rubric-grounded rollout scaffolding, and rubric-based reward modeling~\citep{he2026advancedif, li2026rubrichub, DBLP:journals/corr/abs-2508-12790, zhou2025breaking, zhang2026simple}. Structured rubrics further organize evaluation across heterogeneous criteria~\citep{starace2025paperbench, condor2022representing}. These works primarily focus on rubric design and use, leaving the aggregation of fixed, heterogeneous criterion outputs into a scalar policy-optimization signal largely unaddressed.

\paragraph{Criterion-level reward aggregation.}
Existing methods mainly aggregate criterion scores in cardinal space using weighted sums, normalization, dependency modeling, or adaptive reweighting~\citep{liu2026gdpo, lv2026mitigating, tyagi2026not}. Related work also explores relative reward signals in group-based RL~\citep{niu2026absolute}. However, these approaches largely rely on cardinal score aggregation and therefore require heterogeneous criteria to be normalized or otherwise aligned. They also typically treat auxiliary behavioral attributes as reward criteria, which can alter the objective rather than merely control their systematic influence. Detailed comparisons are provided in Appendix~\ref{app:related}.

\section{Conclusion}

We introduce \textbf{RLVR$^{2}$}, a verifiable ranking paradigm for reinforcement learning with heterogeneous criteria. Across three model scales and 20 benchmarks, RLVR$^{2}$ consistently achieves leading performance while supporting attribute control and adaptive weighting. Future work will extend RLVR$^{2}$ to broader tasks, verifiers, and ranking models.



\subsection*{Ethics statement}

This work studies reward fusion for RL post-training. It involves no human subjects or human annotation, and all benchmarks are public research datasets used under their respective licenses. Outputs from the LLM judges may be incorrect or unsafe; such content does not reflect the authors' views.

\clearpage
\bibliography{Main}
\bibliographystyle{colm2024_conference}

\clearpage
\appendix
\section{Dataset and Benchmark}
\label{appendix:dataset}

In this section, we discuss the training dataset and evaluation dataset used in our experiments.

\begin{table}[htbp]
  \centering
  \begin{subtable}[t]{0.48\linewidth}
    \centering
    \footnotesize
    \setlength{\tabcolsep}{6pt}
    \caption{General-chat training mixture.}
    \begin{tabular}{@{}lr@{}}
      \toprule
      Domain & \# Queries \\
      \midrule
      Instruction following & 24{,}900 \\
      Open-ended generation & 11{,}881 \\
      Factuality \& QA & 8{,}819 \\
      Long-context understanding & 8{,}100 \\
      Multi-turn dialogue & 2{,}400 \\
      \midrule
      \textbf{Total} & \textbf{56{,}100} \\
      \bottomrule
    \end{tabular}
    \label{tab:general_train_data}
  \end{subtable}\hfill
  \begin{subtable}[t]{0.48\linewidth}
    \centering
    \footnotesize
    \setlength{\tabcolsep}{6pt}
    \caption{Agentic training mixture.}
    \begin{tabular}{@{}lr@{}}
      \toprule
      Task family & \# Queries \\
      \midrule
      Software engineering & 25{,}591 \\
      Reasoning & 18{,}515 \\
      Data-analysis agents & 15{,}289 \\
      Function-calling agents & 3{,}618 \\
      Computer-use/coding agents & 718 \\
      Domain-specific agents & 389 \\
      \midrule
      \textbf{Total} & \textbf{64{,}120} \\
      \bottomrule
    \end{tabular}
    \label{tab:agentic_train_data}
  \end{subtable}
  \caption{Composition of the two-stage training data. Internal data sources are grouped into broad domains and task families.}
  \label{tab:train_data}
\end{table}

\subsection{Training Dataset Statistics}
\label{appendix:train_dataset}

Table~\ref{tab:general_train_data} summarizes the composition of our 56.1k-query
training mixture. The data is organized into five broad domains that jointly
cover the six capability dimensions evaluated in Section~\ref{sec:setup}:
\emph{instruction following}, \emph{open-ended chat and generation},
\emph{factuality and QA}, \emph{long-context}, and \emph{multi-turn dialogue}.
Instruction following is the largest domain (24.9k, roughly 44\% of the
mixture), reflecting our emphasis on precise constraint satisfaction and its
central role in aligning model behavior with user intent; it spans a diverse
set of constraint types, including format, length, conditional, and
cross-lingual instructions, in both English and Chinese. The remaining domains
balance open-ended generation, short- and long-form factual QA, long-context
comprehension, and multi-turn consistency, so that the policy is exposed to
both rule-verifiable and open-ended tasks during training. Each query is
trained with a maximum decode length of 16{,}384 tokens. Each query activates a task-dependent subset of the shared rubric library, producing binary, discrete, or continuous criterion scores. All compared aggregation methods use the same training queries, active criteria, scoring procedures, and judges; they differ in how the resulting criterion scores are synthesized for policy optimization. To
preserve anonymity, internal data sources are referred to by generic
capability labels rather than product names.

\subsection{Agentic Training Dataset}
\label{appendix:agentic_train_dataset}

For the agentic experiments, we use an additional mixture of 64{,}120 queries spanning software engineering, reasoning, data analysis, function calling, computer-use and coding agents, and other domain-specific agent tasks. Table~\ref{tab:agentic_train_data} reports its composition. Software-engineering tasks form the largest component and include repository-level problem solving, code repair, code understanding, and related coding-agent settings. Data-analysis tasks require agents to inspect files and structured data using tools, while function-calling tasks evaluate selecting and invoking external functions. The reasoning component broadens the mixture with logic, science, mathematics, and general problem-solving queries. Every example includes a verifier and a length-penalty configuration; tool definitions are provided when required by the task. All compared methods use the same queries, tools, verifiers, and training settings, differing only in reward construction.

\section{Criterion list}
\label{sec:criterion_list}

This section catalogs representative criteria from the rubric library used in our experiments. The library contains more than 180 criteria, with task-dependent subsets activated for each query. Each criterion is given a descriptive name, a one-line description, its semantic family, and its output value domain: Binary $\{0,1\}$, Ternary $\{0,0.5,1\}$, Continuous $[0,1]$, or Multi (other discrete sets, e.g.\ $\{-1,0,1\}$). Criteria are split into two groups: \emph{base} criteria emit positive quality signals (Table~\ref{tab:crit_base}), while \emph{penalty} criteria detect low-quality behavior and subtract from the score (Table~\ref{tab:crit_low}).

\newcolumntype{Y}{>{\raggedright\arraybackslash}X}

\begin{table}[htbp]
\centering\footnotesize
\caption{Base criteria (positive quality signals), grouped by semantic family.}
\label{tab:crit_base}
\begin{tabularx}{\linewidth}{l Y c}
\toprule
\textbf{Criterion} & \textbf{Description} & \textbf{Value} \\
\midrule
\multicolumn{3}{l}{\textit{Comprehensive content quality}} \\
GeneralQuality           & LLM-based overall quality score & Continuous \\
BusinessScenarioQuality  & Holistic quality in a business scenario & Multi \\
CoreIntentMatch          & Core-intent understanding \& relevance & Binary \\
TaskTypeMatch            & Correctness of task-type recognition & Binary \\
\addlinespace

\multicolumn{3}{l}{\textit{Content faithfulness \& memory}} \\
CoreFactAccuracy         & Coverage/correctness of core facts & Ternary \\
NonCoreFactAccuracy      & Coverage/correctness of non-core facts & Ternary \\
MaterialFaithfulness     & Faithfulness to provided source material & Binary \\
MultiTurnMemory          & Context consistency across dialogue turns & Binary \\
\addlinespace

\multicolumn{3}{l}{\textit{Format compliance}} \\
GeneralFormatCompliance  & General structural/formatting compliance & Binary \\
FormatRequirement        & Explicit format-requirement satisfaction & Binary \\
TextTypeMatch            & Correct text type/genre (email, article) & Binary \\
BulletCount              & Number of bullet-list items & Binary \\
HighlightCount           & Number of highlighted sections & Binary \\
ParagraphCount           & Number of paragraphs & Binary \\
CaseFormat               & Upper/lower-case formatting requirement & Binary \\
TitleFormat              & Title presence and format & Binary \\
QuotationFormat          & Quotation formatting requirement & Binary \\
Postscript               & Required postscript section & Binary \\
PromptRepeat             & Must restate the original prompt first & Binary \\
ConstrainedResponse      & Response limited to a fixed option set & Binary \\
ConstrainedWriting       & Constrained creative writing (acrostic) & Multi \\
SummaryRecall            & Recall-based summary coverage & Continuous \\
\addlinespace

\multicolumn{3}{l}{\textit{Third-party benchmark detectors}} \\
IFBench                  & IFBench format/logic check & Multi \\
MultiIF                  & Multi-IF benchmark check & Multi \\
Collie                   & Collie benchmark check & Multi \\
ComplexBench             & ComplexBench benchmark check & Continuous \\
LiveBench                & LiveBench benchmark check & Continuous \\
\addlinespace

\multicolumn{3}{l}{\textit{Length \& counting}} \\
WordCount                & Word/char count constraint & Binary \\
SentenceCount            & Sentence-count constraint & Binary \\
EntryCount               & Number of entries & Binary \\
LetterFrequency          & English letter-frequency constraint & Binary \\
\addlinespace

\multicolumn{3}{l}{\textit{Language \& locale}} \\
LanguageRequirement      & Required output language & Binary \\
LanguageDetection        & Detected language of the response & Binary \\
CrossLingualInstruction  & Cross-lingual instruction compliance & Binary \\
\addlinespace

\multicolumn{3}{l}{\textit{Style \& wording}} \\
DialogueStyle            & Consistency of dialogue style & Binary \\
DictionRequirement       & Diction / wording-norm requirement & Binary \\
\addlinespace

\multicolumn{3}{l}{\textit{Identity \& capability boundary}} \\
RoleConsistency          & Identity/role consistency requirement & Binary \\
CapabilityBoundary       & Capability-boundary / identity limits & Binary \\
\addlinespace

\multicolumn{3}{l}{\textit{Content presence \& matching}} \\
Keyword                  & Required keyword presence & Binary \\
ForbiddenContent         & Forbidden word/content detection & Binary \\
GroundTruthMatch         & Match against a gold answer & Binary \\
\bottomrule
\end{tabularx}
\end{table}

\begin{table}[htbp]
\centering\footnotesize
\caption{Penalty criteria (low-quality signals that subtract from the score).}
\label{tab:crit_low}
\begin{tabularx}{\linewidth}{l Y c}
\toprule
\textbf{Criterion} & \textbf{Description} & \textbf{Value} \\
\midrule
GeneralLowQuality        & Perfunctory / template-like output & Binary \\
RoboticTemplate          & Robotic/templated output detection & Binary \\
Verbosity                & Redundancy / conciseness penalty & Binary \\
Affectation              & Affected / overwrought wording & Binary \\
ImproperWording          & Improper wording detection & Binary \\
LogicalError             & Logical-error detection & Binary \\
UnrealisticContent       & Unrealistic / non-factual content & Binary \\
SecurityRisk             & Security-risk detection & Binary \\
\bottomrule
\end{tabularx}
\end{table}

\subsection{Evaluation Benchmarks Statistics}
\label{appendix:eval_dataset}

\textbf{SysBench.} SysBench \citep{qin2024sysbench} evaluates whether large
language models can reliably follow system messages. It assesses models along constraint following, instruction misalignment, and multi-turn stability, targeting the gap between stated system-level requirements and actual model behavior.

\textbf{AdvancedIF.} AdvancedIF \citep{he2026advancedif} is a rubric-based
benchmark for advanced instruction following. Each prompt is paired with a set of expert-written rubric criteria, enabling fine-grained evaluation of whether a response satisfies complex, compositional instructions beyond surface-level compliance.

\textbf{IFEval.} IFEval \citep{zhou2023instruction} measures instruction
following through automatically verifiable instructions, such as constraints on output length, format, or the inclusion of specified keywords. Because compliance is checked programmatically rather than by a judge, the benchmark provides an objective, low-noise signal of instruction adherence.

\textbf{IFBench.} IFBench \citep{pyatkin2026generalizing} extends verifiable instruction following to a broader and more diverse set of constraints, emphasizing generalization to instruction types not seen during training. It is designed to test whether models follow the underlying intent of a constraint rather than overfitting to a fixed instruction template.

\textbf{MultiChallenge.} MultiChallenge \citep{deshpande2025multichallenge} is a realistic multi-turn conversation benchmark that is challenging for frontier LLMs. It evaluates a model's ability to maintain instruction adherence and contextual consistency across multiple dialogue turns, where earlier constraints must continue to hold in later responses.


\textbf{LongBench-v2.} LongBench-v2 \citep{bai2025longbench} targets deeper
understanding and reasoning over realistic long-context multitasks. It spans a range of long-document scenarios and requires genuine comprehension and multi-step reasoning rather than simple retrieval from the context.

\textbf{LongProc.} LongProc \citep{ye2025longproc} benchmarks long-context language models on long procedural generation, where a model must produce lengthy, structured outputs by following multi-step procedures. It stresses both long-context comprehension and long-form generation consistency.

\textbf{HELMET.} HELMET \citep{yen2025helmet} is a benchmark for evaluating long-context models effectively and thoroughly across diverse application-driven categories. It is designed to provide a more reliable and comprehensive picture of long-context capability than single-task or synthetic evaluations.

\textbf{AA-LCR.} The Artificial Analysis Long Context Reasoning benchmark (AA-LCR) \citep{artificialanalysis2025lcr} evaluates reasoning over long inputs, requiring models to integrate and reason over information distributed across an extended context rather than locate a single passage.

\textbf{SimpleQA-verified.} SimpleQA-verified \citep{haas2025simpleqa} is a
reliable factuality benchmark that measures a model's parametric knowledge
through short, fact-seeking questions with verified reference answers. Its
verified answer set is intended to reduce label noise and provide a trustworthy factuality signal.


\textbf{ChineseSimpleQA.} ChineseSimpleQA \citep{he2025chinese} is a Chinese factuality evaluation for large language models, consisting of short fact-seeking questions in Chinese. It provides a factuality signal
complementary to the English benchmarks and covers a broad range of knowledge domains.

\textbf{Frames.} Frames \citep{krishna2024factfetchreasonunified} provides a unified evaluation of retrieval-augmented generation, jointly assessing factuality, retrieval, and reasoning. It requires models to fetch relevant
evidence and reason over it to arrive at correct answers, rather than testing these abilities in isolation.

\textbf{GPQA.} GPQA \citep{rein2023gpqa} is a graduate-level, ``Google-proof'' multiple-choice question-answering benchmark spanning biology, chemistry, and physics, with questions ranging from advanced undergraduate to postgraduate difficulty. The questions are constructed so that they cannot be answered reliably through simple web search, testing deep domain reasoning.

\textbf{MMLU-Pro.} MMLU-Pro \citep{wang2024mmlu} is a more robust and
challenging multi-task language understanding benchmark. It extends MMLU with harder, reasoning-intensive questions and an expanded answer-option set, reducing the effect of guessing and providing a stronger measure of broad knowledge.

\textbf{ArenaHard-v2.} ArenaHard-v2 \citep{li2024crowdsourced} is a high-quality open-ended benchmark constructed from crowdsourced conversational data via the Arena-Hard / BenchBuilder pipeline. It consists of challenging open-ended prompts and is designed to correlate well with human preference on real user queries.

\textbf{EQBench3.} EQBench3 \citep{paech2023eqbench} evaluates emotional
intelligence in open-ended dialogue, scoring responses on their handling of
interpersonal and affective content in multi-turn role-play scenarios rather
than on task correctness. It complements ArenaHard-v2 by probing a dimension of
open-ended quality that correctness-oriented judges do not capture.

\textbf{$\tau^3$-Bench Banking.} $\tau^3$-Bench Banking \citep{shi2026tau} evaluates agents in realistic multi-turn banking scenarios, where an agent must interact with a user, invoke domain tools, and satisfy task-specific policies to complete a request. It tests reliable tool use and task completion under evolving conversational state.

\textbf{GAIA.} GAIA \citep{mialon2024gaia} evaluates general-purpose assistants on real-world questions that require combinations of reasoning, web or file-based information retrieval, and tool use. Its multi-step tasks measure whether an agent can coordinate these capabilities to produce a verifiable final answer.

\textbf{GDPVal.} GDPVal \citep{patwardhan2026gdpval} evaluates AI systems on economically valuable knowledge-work tasks drawn from professional domains. It emphasizes the quality of complete task deliverables, providing a complementary measure of agent performance beyond short-answer reasoning and isolated tool calls.

\textbf{SWE-bench Verified.} SWE-bench Verified \citep{jimenez2024swebench} evaluates repository-level software engineering by asking an agent to resolve validated issues in real-world codebases. Success requires understanding the repository, localizing the fault, modifying code, and producing a patch that passes the associated tests.

\begin{table*}[t]
  \centering
  \tiny
  \setlength{\tabcolsep}{2.4pt}
  \renewcommand{\arraystretch}{1.15}
  \resizebox{\textwidth}{!}{%
  \begin{tabular}{cll*{14}{c}}
    \toprule
    & & & \multicolumn{4}{c}{\textbf{Instruction Following}}
      & \multicolumn{2}{c}{\textbf{Factuality \& QA}}
      & \multicolumn{4}{c}{\textbf{Long-Context}}
      & \multicolumn{2}{c}{\textbf{Reason.}}
      & \multicolumn{1}{c}{\textbf{Know.}}
      & \multicolumn{1}{c}{\textbf{Open}} \\
    \cmidrule(lr){4-7}\cmidrule(lr){8-9}\cmidrule(lr){10-13}\cmidrule(lr){14-15}\cmidrule(lr){16-16}\cmidrule(l){17-17}
    & \textbf{Method} & \textbf{Compr.}
      & SysB & AdvIF & IFE & IFB
      & SQA-v & CSQA
      & LB-v2 & LProc & HLM & Frm
      & AA-L & GPQA
      & MMLU & EQB3 \\
    \midrule
    \multirow{6}{*}{\emph{9B}}
      & Rule-Fusion & \ding{55}
        & $74.12$ & $62.23$ & $90.76$ & $48.59$
        & $11.60$ & $67.13$
        & $32.01$ & $81.75$ & $25.86$ & $74.64$
        & $55.33$ & $80.68$
        & $80.39$
        & $86.65$ \\
    \cmidrule(l){2-17}
      & RLVR$^{2}$+ & \ding{55}
        & $76.44$ & $65.56$ & $92.61$ & $73.81$
        & $12.20$ & $69.50$
        & $32.21$ & $82.55$ & $26.00$ & $76.21$
        & $60.00$ & $81.94$
        & $81.11$
        & $85.65$ \\
      & RLVR$^{2}$+ & \emph{on}
        & $73.80$ & $64.95$ & $92.05$ & $72.11$
        & $11.83$ & $68.91$
        & $32.21$ & $81.14$ & $26.20$ & $75.36$
        & $61.33$ & $80.68$
        & $80.85$
        & $85.85$ \\
      & RLVR$^{2}$+ & $0.25$
        & $74.04$ & $63.50$ & $91.13$ & $72.45$
        & $14.11$ & $70.63$
        & $32.80$ & $81.76$ & $26.47$ & $75.24$
        & $57.33$ & $83.08$
        & $80.72$
        & $87.30$ \\
      & RLVR$^{2}$+ & $0.50$
        & $75.40$ & $65.29$ & $91.68$ & $70.75$
        & $14.19$ & $69.31$
        & $33.00$ & $81.64$ & $23.51$ & $76.09$
        & $59.00$ & $81.19$
        & $81.41$
        & $85.30$ \\
      & RLVR$^{2}$+ & $0.75$
        & $74.60$ & $66.69$ & $92.42$ & $77.21$
        & $13.36$ & $70.36$
        & $32.80$ & $83.10$ & $26.03$ & $75.73$
        & $59.67$ & $81.31$
        & $81.31$
        & $85.60$ \\
    \bottomrule
  \end{tabular}%
  }
  \caption{\textbf{Effect of thinking-efficiency compression.} All rows use the same 9B backbone with settings otherwise fixed. MultiChallenge and ArenaHard-v2 are omitted. Judge models and scoring protocols are given in Appendix~\ref{appendix:eval_dataset}.}
  \label{tab:appendix_think_compr}
\end{table*}

\section{Rollout Size Detail Impact}
\label{app:gropu_size_result}

We report per-benchmark results for the rollout-group-size ablation. All settings other than the group size are held fixed.

\begin{table*}[htbp]
  \centering
  \tiny
  \setlength{\tabcolsep}{2.4pt}
  \renewcommand{\arraystretch}{1.15}
  \resizebox{\textwidth}{!}{
  \begin{tabular}{clcccccccccccccccc}
    \toprule
    & & \multicolumn{5}{c}{\textbf{Instruction Following}}
      & \multicolumn{2}{c}{\textbf{Factuality \& QA}}
      & \multicolumn{4}{c}{\textbf{Long-Context}}
      & \multicolumn{2}{c}{\textbf{Reason.}}
      & \multicolumn{1}{c}{\textbf{Know.}}
      & \multicolumn{2}{c}{\textbf{Open-ended}} \\
    \cmidrule(lr){3-7}\cmidrule(lr){8-9}\cmidrule(lr){10-13}\cmidrule(lr){14-15}\cmidrule(lr){16-16}\cmidrule(l){17-18}
    & \textbf{Rollout Size}
      & SysB & AdvIF & IFE & IFB & MChal
      & SQA-v & CSQA
      & LB-v2 & LProc & HLM & Frm
      & AA-L & GPQA
      & MMLU & AH-v2 & EQB3 \\
    \midrule
    & 4 & $73.76$ & $62.67$ & $91.68$ & $71.49$ & $68.33$ & $10.8$ & $66.73$ & $33.20$ & $82.21$ & $25.80$ & $76.09$ & $57.67$ & $82.83$ & $80.67$ & $49.60$ & $85.75$ \\
    & 8 & $76.88$ & $62.70$ & $90.39$ & $68.71$ & $68.56$ & $14.4$ & $65.67$ & $31.81$ & $80.30$ & $26.52$ & $75.36$ & $61.67$ & $81.19$ & $81.28$ & $49.30$ & $86.30$ \\
    & 16 & $73.64$ & $61.70$ & $91.87$ & $71.41$ & $68.73$ & $11.4$ & $66.67$ & $32.01$ & $80.00$ & $26.57$ & $75.12$ & $57.00$ & $79.80$ & $80.98$ & $51.30$ & $85.35$ \\
    \bottomrule
  \end{tabular}
  }
  \caption{Comparison of rollout group size. Judge models and scoring protocols are described in Appendix~\ref{appendix:eval_dataset}.}
  \label{tab:appendix_rollout_size}
\end{table*}

\subsection{Score-Representation Robustness}
\label{app:score-robustness}

We conduct a controlled robustness experiment using real rollout data from the baseline run. The comparison replays multiple aggregation rules on exactly the same response groups, avoiding differences caused by distinct sampled responses. We retain groups with exactly $G=8$ unique responses and require every retained criterion to have a finite score in $[0,1]$ for all responses in the group. Quality-gate verifiers and aggregation-derived fields are excluded from the criterion matrix, including \texttt{RepeatSegmentVerifier}, \texttt{focal\_score}, \texttt{gdpo\_prenorm}, \texttt{gdpo\_reward}, \texttt{original\_reward}, and \texttt{sc\_fused\_score}.

For each retained group, we construct a criterion matrix from the original scores and replay each aggregation rule before and after transforming one target criterion. The transformations preserve within-criterion ordering while changing numerical spacing. Specifically, we multiply IFBench scores by $10$, remap the UniLLM score levels from $(0,0.2,1)$ to $(0,1,100)$, and square the UniBusiness scores.
We compare raw cardinal summation, group-wise $z$-score normalization, and RLVR$^{2}$. For RLVR$^{2}$, every criterion is converted into pairwise win/tie/loss outcomes within the group, and a Bradley--Terry utility is fitted from these outcomes before criterion-level fusion. Thus, the experiment evaluates the stability of the final fused signal rather than only the stability of an individual criterion ranking.

\begin{table*}[t]
  \centering
  \footnotesize
  \setlength{\tabcolsep}{6pt}
  \renewcommand{\arraystretch}{1.12}
  \caption{\textbf{Robustness to order-preserving score transformations.} Results are computed on identical $G=8$ response groups before and after transforming one criterion. Higher values indicate greater stability.}
  \label{tab:score-robustness}
  \begin{tabular}{@{}llrrr@{}}
    \toprule
    Criterion & Aggregation & Groups & Sign agreement (\%) & Spearman $\rho$ \\
    \midrule
    IFBench & Cardinal & 672 & 92.52 & 0.9783 \\
             & Normalized cardinal & 672 & 98.75 & 0.9985 \\
             & RLVR$^{2}$--BT & 672 & \textbf{100.00} & \textbf{1.0000} \\
    \midrule
    UniLLM & Cardinal & 194 & 95.88 & 0.9651 \\
            & Normalized cardinal & 194 & 99.16 & 0.9921 \\
            & RLVR$^{2}$--BT & 194 & \textbf{100.00} & \textbf{1.0000} \\
    \midrule
    UniBusiness & Cardinal & 954 & 95.83 & 0.99997 \\
                & Normalized cardinal & 954 & 97.68 & 0.99925 \\
                & RLVR$^{2}$--BT & 954 & \textbf{100.00} & \textbf{1.0000} \\
    \bottomrule
  \end{tabular}
\end{table*}

Here, ``Sign agr.'' is the fraction of responses whose sign relative to the group-mean fused reward is unchanged before and after transformation. Spearman correlation measures agreement of the resulting within-group reward ordering. The invariance result is conditional on the transformation preserving the pairwise outcome matrix, including any tie decisions. With a nonzero tie margin, a numerical transformation can change which pairs are treated as ties; such transformations are outside the strict invariance claim.

\section{Extended Related Work}
\label{app:related}

Open-ended generation is commonly evaluated through multidimensional LLM-as-a-judge protocols \citep{zheng2023judging,liu2023g,gu2026survey}. Rubrics make these judgments interpretable by decomposing response quality into fine-grained criteria \citep{kim2024prometheus,arora2025healthbench}. Rubric-guided RL subsequently uses such criteria in several forms, including instruction-following pipelines \citep{he2026advancedif}, automatically scaled datasets \citep{li2026rubrichub}, reward anchors \citep{DBLP:journals/corr/abs-2508-12790}, rollout scaffolds \citep{zhou2025breaking}, and in-context specifications \citep{zhang2026simple}. RLVR$^{2}$ complements these approaches by focusing on the downstream aggregation of heterogeneous criterion-level judgments into a scalar training signal.

Existing aggregation methods span flat, structured, and adaptive formulations. Weighted sums treat rubric items as independent components \citep{li2026rubrichub}, while hierarchical subtasks and rubric graphs organize dependencies among criteria \citep{starace2025paperbench,condor2022representing}. Group reward-decoupled normalization and policy-aware reweighting adapt the contribution of cardinal reward components \citep{liu2026gdpo,tyagi2026not}. GEAR instead propagates soft suppression over a query-specific rubric graph to prevent unsupported criteria from contributing utility \citep{lv2026mitigating}. RLVR$^{2}$ occupies a complementary position: it aggregates criterion-wise ordinal comparisons and applies explicit quality-level gating, combining reduced dependence on cross-criterion calibration with non-compensatory quality control.


\end{document}